\documentclass[10pt,twocolumn,letterpaper]{article}

\usepackage{wacv}              

\usepackage{graphicx}
\usepackage{amsmath}
\usepackage{amssymb}
\usepackage{booktabs}
\usepackage{comment}

\usepackage{multirow} 

\usepackage[pagebackref,breaklinks,colorlinks]{hyperref}
\newcommand{\rr}[1]{\rotatebox{45}{#1}}

\usepackage[capitalize]{cleveref}
\crefname{section}{Sec.}{Secs.}
\Crefname{section}{Section}{Sections}
\Crefname{table}{Table}{Tables}
\crefname{table}{Tab.}{Tabs.}

\def\wacvPaperID{1358} 
\def\confName{WACV}
\def\confYear{2025}

\begin{document}

\title{Optical Flow Matters: an Empirical Comparative Study \\
on Fusing Monocular Extracted Modalities for Better Steering}

\author{Fouad Makiyeh$^{1}$, Mark Bastourous$^{1}$, Anass Bairouk$^{1}$, Wei Xiao$^{2}$, Mirjana Maras$^{1}$, Tsun-Hsuan Wangb$^{2}$, \\
Marc Blanchon$^{1}$, Ramin Hasani$^{2}$, Patrick Chareyre$^{1}$ and Daniela Rus$^{2}$ \\
$^{1}$Hybrid Intelligence part of Capgemini Engineering\\
$^{2}$Computer Science and Artiﬁcial Intelligence Lab, Massachusetts Institute of Technology}
\maketitle

\begin{abstract}
Autonomous vehicle navigation is a key challenge in artificial intelligence, requiring robust and accurate decision-making processes. This research introduces a new end-to-end method that exploits multimodal information from a single monocular camera to improve the steering predictions for self-driving cars. Unlike conventional models that
require several sensors which can be costly and complex or rely exclusively on RGB images that may not be robust enough under different conditions, our model significantly improves vehicle steering prediction performance from a single visual sensor. 
By focusing on the fusion of RGB imagery with depth completion information or optical flow data, we propose a comprehensive framework that integrates these modalities through both early and hybrid fusion techniques. 

We use three distinct neural network models to implement our approach: Convolution Neural Network - Neutral Circuit Policy (CNN-NCP) , Variational Auto Encoder - Long Short-Term Memory (VAE-LSTM) , and Neural Circuit Policy architecture VAE-NCP. 
By incorporating optical flow into the decision-making process,  our method significantly advances autonomous navigation. Empirical results from our comparative study using Boston driving data show that our model, which integrates image and motion information, is robust and reliable. It outperforms  state-of-the-art approaches that do not use optical flow, reducing the steering estimation error by 31\%. This demonstrates the potential of optical flow data, combined with advanced neural network architectures (a CNN-based structure for fusing data and a Recurrence-based network for inferring a command from latent space), to enhance the performance of autonomous vehicles steering estimation.
\end{abstract}

\begin{figure}[ht]
    \centering    \includegraphics[width=0.48\textwidth, keepaspectratio]{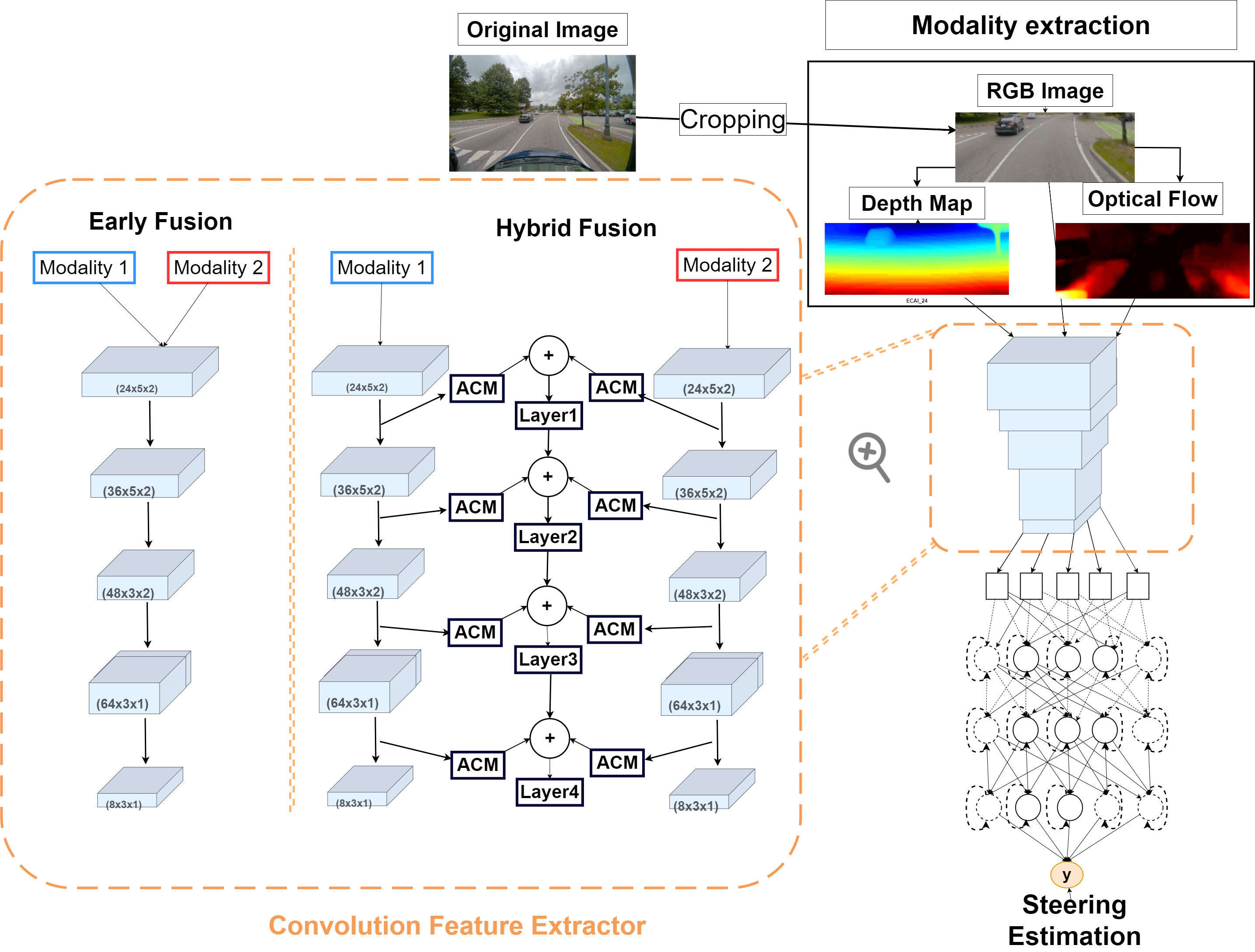} 
    \caption{Block diagram of the overall approach. An example of different modalities extracted from an RGB image, such as the depth map and optical flow. Two of these modalities are fused in a convolutional feature extractor followed by a recurrent neural network for vehicle steering estimation. }
    \label{fig:block_diagram}
\end{figure}

\section{Introduction}
\label{sec:intro}
Self-driving cars play a critical role in improving transportation  by enhancing accessibility, efficiency and safety. These cars have intricate systems that enable them to perceive their environment and produce precise control movements, but they can sometimes take unsafe routes through conflict areas~\cite{schwarting2018planning}. Reaching full autonomy requires expertise in multiple areas: perception, localization, planning, and control~\cite{singh2023endtoend}. 
In this work, we focus on the steering estimation, a vital aspect of control that keeps vehicles within road bounds so that they can safely move around. Challenge arises with varying road conditions such as different road colors, textures, lighting and lane markings ~\cite{dong2023applications,Tampuu2020EndToEndDriving}.

Traditional methods of steering estimation use modular systems that deal with perception decision-making and control separately. However, these methods tend to be effective but have complex systems which need exact calibration as well as integration of different components possibly leading to mistakes as well as inefficiencies~\cite{chen2015deepdriving}. Recently end-to-end deep learning has emerged, where neural networks can be employed  to learn vehicle steering directly from raw input data, such as images gathered by on-board cameras, offering a streamlined approach without needing a full vehicle model~\cite{kim2017interpretable, xu2017end, Lechner2020, bairouk2024}. Due to the effectiveness of such strategies in steering estimation, we use them in our work. Nonetheless, interpretability of such models has been one of the obstacles for the successful deployment in self-driving cars. 
An investigation into the interpretability and robustness of our method will be done using a new technique introduced in~\cite{bairouk2024}.

Moreover, conventional supervised end-to-end lateral control approaches have typically relied on RGB images as primary data sources, as presented in~\cite{codevilla2019exploring, ohn2020learning, prakash2020exploring}. While these frame-based approaches have typically solely relied on RGB images, the incorporation of multiple modalities, denoted as multimodal fusion, has emerged for addressing limitations of individual sensors~\cite{li2022emergent}. 
Despite extensive investigation into multimodal fusion, the predominant emphasis has been on integrating data from diverse sensors (RGB camera, Radar, Lidar)~\cite{li2022emergent}, thereby neglecting the exploration of methods leveraging the abundance of visual data obtained from a single camera. Moreover, this process of integration from multiple sensors is computationally intensive and requires significant processing capability to help interpret data from many sensors in real time. This leads to higher power consumption which is a major concern for electric vehicles. Furthermore, sensor fusion systems must also address issues related to data synchronization, calibration, and the potential for conflicting information from different sensors, which can complicate the decision-making process~\cite{liu2020computing}.

This study aims to explore the potential of deriving multiple modalities from a single camera and leveraging deep neural network techniques to fuse these modalities for enhanced prediction of vehicle steering angles. 
The novelty of our work lies in enhancing steering estimation using modalities extracted from a single visual sensor and advanced networks like Neural Control Policy (NCP) network, which has demonstrated its efficiency in such applications~\cite{Lechner2020}. Unlike NCP-based approaches that use solely RGB images, our approach employs this network with multiple modalities. 
The primary contributions of our research are then summarized as follows:
\begin{itemize}
    \item We investigate various modalities extractable from RGB images, with a particular emphasis on depth maps and optical flow images. Subsequently, we assess the effectiveness of different fusion techniques within the context of vehicle steering estimation.  
    \item Our study demonstrates substantial improvements in three different end-to-end vehicle steering estimation architectures - CNN-NCP, VAE-LSTM and VAE-NCP - through the integration of optical flow. Our experiments indicate a notable decrease in the mean squared error (MSE) of steering estimation, reducing from 3.17, as achieved solely with RGB images using state-of-the-art (SOTA) approaches, to 1.64 with our proposed method.
    \item We illustrate the augmented generalization and robustness of our model, highlighting the significant advantages resulting from the inclusion of optical flow information within the predictive framework.
\end{itemize}

\section{Related works}
Artu{n}edo \etal\cite{artunedo2024lateral} presented a comprehensive comparison between model-based and model-free approaches to assess the effectiveness of different lateral control algorithms used in autonomous driving systems.
Model-based control strategies use a mathematical model of the vehicle dynamics, while model-free strategies use data-driven methods, both to generate appropriate control actions.
Because of the dynamic problem and complicated nature, we will focus on and employ a model-free approach, utilizing deep learning techniques explicitly to benefit from their data-driven learning abilities and flexibility when dealing with diverse situations without the requirement of set models.

Bojarski \etal~\cite{bojarski2016} developed a vision-based end-to-end driving system based on deep CNNs that directly mapped raw pixel values from a single front camera to steering commands. In this way, they trained their model on lots of driving footage data sets which in turn made it possible for the model to accommodate variety of traffic conditions. Nevertheless, as far as more complex driving behaviors such as lane changes their scope was limited. Instead of using RGB images as input, Linda \etal~\cite{capito2020optical} exploited optical flow, the motion of objects between frames, to dynamically perceive the environment and generate a visual potential field whereby the direction and magnitude of flow vectors affects navigation decisions of the vehicle. By including optical flow in decision making, the system capability to navigate complicated driving scenarios was enhanced.

Other approaches focused on including time series instead of conventional CNNs. For instance, 
Eraqi \etal~\cite{eraqi2017end} aimed at improving the steering angle control in a car through using Long Short Term Memory (LSTM).
The method was instrumental in enabling the system to have an enhanced knowledge of time context in driving conditions, making steering predictions to be more precise and stable over time. Similarly, Xu \etal~\cite{xu2017end}  created a model for driving by using fully convolutional network with LSTM for predicting steering angle. The authors included semantic segmentation as an auxiliary task to improve the model perception of the road conditions and consequently better predict steering angles. Additionally, Lechner \etal~\cite{Lechner2020} introduced Neural Circuit Policies (NCPs), a 
brain-inspired neural computation method that provides interpretable and accurate decision-making maps from high-dimensional inputs.
LSTM and NCP provide robust steering estimation results, particularly under varying road illumination conditions, making them strong candidates for consideration in our work.

To further enhance perception and decision-making capabilities, integrating multiple sensor modalities has been explored. Fusing RGB images with Lidar data, radar images and  depth maps improved several perception tasks such as object recognition, tracking and semantic segmentation~\cite{Fooladgar2019MultiModalAF, xiao2020multimodal, schneider2017multimodal}. Motivated by this, we present different studies that examine multimodality in end-to-end networks for steering estimation.  There are several fusion methods, such as early fusion (concatenating raw data from different modalities into multiple input channels), late fusion (concatenating multimodal feature maps at the decision level), and hybrid fusion (fusing features at intermediate layers)~\cite{feng2020deep}. High image calibration and synchronization are needed for early fusion, late fusion learns unimodal features independently but may lack cross-modal correlation, while hybrid fusion combines the strengths of both. Munir \etal~\cite{munir2022multimodal} addressed the challenge of predicting steering angles by integrating data from event-based cameras and frame-based RGB cameras. 
The hybrid fusion of frame-based RGB data and event camera data was done through the use of self-attention layers between the convolutional encoders for each modality. This self-attention mechanism helps in learning long term relationships therefore, allowing the model to combine features from both sources before being passed to the decoder which predicts steering angle. Additionally,  they showed that integrating event cameras leads to enhanced steering angle prediction. 
Maanp\"a\"a et al. \cite{maanpaa2021multimodal} used RGB images and Lidar range and reflectance data to forecast steering angles in challenging weather conditions. They adopted two fusion methodologies: the middle fusion dual model and the channel gated dual model. In the middle fusion strategy, features are fused after the convolutional phase of the network, whereas the channel gated dual model consists of two parallel architectures that integrate features through gated channels.
Abou-Hussein et al. \cite{abou2019multimodal} conducted a series of experiments to evaluate the efficacy of spatial, spatiotemporal, and multimodal approaches in predicting automotive steering. 
Various strategies were tested, including a standard CNN, a 3D-CNN stacking three sequential grayscale images, spatial + difference method incorporating a difference frame between current and previous frames, and spatial with optical flow technique involving normalization of optical flow between current and previous frames. 
Upon comparison with adopting of a Recurrent Neural Network (RNN), it was concluded that spatial + optical flow and spatial + difference variations outperformed simply stacking images in the 3D-CNN, emphasizing the advantage of explicitly incorporating temporal information into the model. 
Yi \etal~\cite{xiao2020multimodal} explored combining RGB images and depth data. They evaluated early, mid, and late fusion methods for integrating multimodal data.
They showed that early fusion of multimodal data significantly enhances driving policy performance in diverse conditions. Shuai \etal~\cite{wang2021flowdrivenet} presented an end-to-end network for training driving policies which employs optical flow from images and point flow from LiDAR data. They showed that the dense motion information provided by optical flow combined with accurate spatial measurements from laser enables the net to navigate complex and dynamic environments skillfully. 
They compared their approach with traditional approaches that are either purely visual or LIDAR-based; hence demonstrating how a multmodal perception system benefits us. 
\section{Methodology}
\noindent\textbf{Overview of the Method}. Our method introduces a novel approach to end-to-end steering estimation for autonomous vehicles by leveraging multimodal information from a single monocular camera. Unlike previous methods that rely solely on RGB data or require multiple sensors, our approach fuses RGB imagery with optical flow data extracted from the same source. This method captures both spatial and temporal dynamics of the driving scene, providing a richer representation for steering estimation. Figure \ref{fig:block_diagram} illustrates our approach. The method consists of four main components:
\begin{enumerate}
\item \textit{Modality Extraction:} From the input RGB image sequence, we extract two key modalities Depth and Optical flow, but only consider Optical Flow, as we will discuss later.
\item \textit{Fusion Strategy:} We employ an early fusion approach, concatenating RGB and optical flow data before feeding it into the neural network.
\item \textit{Feature Extraction:} A CNN or VAE processes the fused input to extract relevant features.
\item \textit{Temporal Modeling:} We integrate advanced recurrent architectures, including the innovative NCP and traditional LSTM networks, to process global temporal dynamics.
\end{enumerate}

A key innovation in our approach is the multi-scale temporal feature extraction. By using optical flow between two consecutive images, we capture local temporal features that represent short-term motion dynamics. Complementing this, our recurrent neural network architectures (NCP or LSTM) capture global temporal features across the entire sequence of inputs. This dual temporal modeling allows our system to understand both immediate frame-to-frame changes and longer-term driving patterns, a crucial advancement for robust steering estimation.
This novel approach enables us to extract and utilize rich spatiotemporal information from a single camera source, a significant advancement over existing methods. By fusing RGB and optical flow data, we capture static scene geometry, local motion cues, and global temporal patterns, potentially improving the accuracy and robustness of steering estimation compared to unimodal approaches.

In the following subsections, we detail each component of our method, elucidating the innovations in our design choices and how they contribute to improved steering estimation performance in autonomous driving scenarios.

\subsection{Modality Extraction}
We explore the process of extracting multiple modalities from a single monocular camera to enhance steering estimation. We focus on depth maps, and optical flow, each providing unique information about the driving scene.\\

\noindent\textbf{Depth Map Generation.} Although not used in the final model, we explored depth map generation as a potential additional modality. Two pretrained models were tested directly on our dataset and provided satisfactory depth extraction quality, eliminating the need for retraining them. These models are:
\begin{itemize}
\item \textit{Monodepth2 \cite{godard2019digging} :} This method uses a pre-trained ResNet as an encoder to extract high-dimensional features, enabling depth inference by considering scene context.
\item \textit{MiDaS \cite{ranftl2020towards}:} 
This approach focuses on zero-shot cross-dataset transfer, aiming for robust depth estimation across varied environments without fine-tuning. By incorporating dataset mixing during training, this model learns to adapt to diverse scenes, lighting conditions, and other variations present across different datasets.
\end{itemize}

\noindent\textbf{Optical Flow Computation. }Optical flow captures dynamic motion cues between consecutive frames, potentially enhancing steering estimation precision. We employ the Farneback algorithm for its accuracy and reliability in computing dense optical flow. 
This extraction process results in a multimodal representation of the driving scene, capturing both spatial structure and local motion dynamics. 

\subsection{Fusion Strategies}
We explore different strategies to effectively combine information from multiple modalities, specifically RGB images with either depth maps or optical flow data. We investigate two main fusion approaches: Early Fusion (EF) and Hybrid Fusion (HF).\\

\noindent\textbf{Early Fusion. }Early fusion involves concatenating raw data from different modalities into multiple channels before feeding them into the neural network. In our case, we concatenate the RGB image (3 channels) with either the depth map (1 channel) or optical flow (2 channels), resulting in a 4-channel or 5-channel input, respectively. This combined input is then processed end-to-end by a single network.

The early fusion input can be represented as:
\begin{equation}
x_{EF} = [M_1, M_2],
\end{equation}
where $M_1$ represents the RGB image and $M_2$ represents the additional modality (depth or optical flow).\\

\noindent\textbf{Hybrid Fusion. }Hybrid fusion represents a compromise between early and late fusion techniques. In this approach, we use separate encoders for different modalities and fuse the features at intermediate layers. This method allows the network to learn cross-modal features with distinct representations at varying depths. We adopt an approach similar to that proposed in \cite{hu2019acnet}, utilizing an Attention Complementary Module (ACM).
The output of the feature extractor model can be described as:

\begin{equation}
z_{HF} = Layer_4 + ACM(G_5(\tilde{M_1})) + ACM(G_5(\tilde{M_2})),
\end{equation}

where:
\begin{equation}
\tilde{M_1} = G_4(G_3(G_2(G_1(M_1)))),
\end{equation}
$G_1, ..., G_5$ represent the layers of the encoder, and ACM is the Attention Complementary Module as shown in Figure~\ref{fig:block_diagram}.
The hybrid fusion approach allows for more flexibility in learning cross-modal relationships while maintaining the ability to extract modality-specific features.

Our experiments compare the effectiveness of these fusion strategies in the context of vehicle steering estimation. The results, presented in later sections, demonstrate the impact of different fusion approaches on the overall performance of the steering estimation model.\\

\subsection{Steering Estimation Architectures}

Our steering estimation approach utilizes a two-stage architecture: a convolutional feature extractor followed by a recurrent neural network for temporal modeling (see Figure~\ref{fig:block_diagram}). We explore different combinations of these components to determine the most effective architecture for our task.\\

\noindent\textbf{Convolutional Feature Extractors. }We investigate two types of convolutional feature extractors, motivated by the work presented in \cite{Lechner2020, bairouk2024}, for demonstrating their efficiencies in steering estimation:

\begin{itemize}
\item \textit{Convolutional Neural Network (CNN):} We employ the CNN architecture as described in \cite{Lechner2020}. This network is designed to efficiently extract spatial features from the input images.

\item \textit{Variational Autoencoder (VAE):} We also explore the use of a VAE encoder, as detailed in \cite{bairouk2024}. The VAE approach allows for capturing underlying data distributions, potentially enhancing the model ability to generalize.

\end{itemize}

Both feature extractors output a low-dimensional feature vector of size 32, as specified in \cite{bairouk2024}.\\

\noindent\textbf{Recurrent Neural Networks. }For temporal modeling, we consider the following two recurrent architectures based on their successful application in optimizing vehicle steering estimation as shown in~\cite{Lechner2020, bairouk2024} and illustrated in Table.~\ref{tab:comparison}:

\begin{itemize}
\item \textit{Neural Circuit Policy (NCP):} We use the NCP with 19 units, as described in \cite{Lechner2020, bairouk2024}. The NCP is inspired by biological neural circuits and has shown promising results in previous steering estimation tasks.

\item \textit{Long Short-Term Memory (LSTM):} We also employ an LSTM network with 19 units. LSTMs are well-known for their ability to capture long-term dependencies in sequential data.

\end{itemize}

\noindent\textbf{End-to-End Model. }Given the ground truth steering of the vehicle, $\hat{\pmb{y}}$,  the end-to-end steering estimation loss, $L(x, \hat{y})$, is given as follows:


\begin{equation}
L(x, \hat{y}) = \beta L_{VAE} + L_{prediction},
\end{equation}

where:

\begin{equation}
L_{VAE} = \lambda_1 L_{recon}(x, \tilde{x}) + \lambda_2 L_{KL}(\mu, \sigma),
\end{equation}

\begin{equation}
L_{prediction} = \frac{\sum_i w(y_i) (\hat{y_i} - y_i)^2}{\sum_i w(y_i)}; y = RNN(z).
\end{equation}

Here, $\pmb{x}$ representing the inputs to the model, $\beta$ distinguishes between using a CNN ($\beta = 0$) and a VAE ($\beta = 1$) for feature extraction. $\lambda_1 = 0.15$ and $\lambda_2 = \lambda_1 e^{-2}$ are regularization parameters. $\tilde{x}$ denotes the reconstruction of $x$, while $\mu$ and $\sigma$ are the parameters used to sample latent variables in the VAE. The function $w(y)$ gives more attention to samples containing road turns and is defined as $w(y) = exp(|y|^\alpha)$.

This architecture allows us to capture both spatial features through the convolutional extractors and temporal dynamics through the recurrent networks, providing a comprehensive approach to steering estimation. The combination of these components enables our model to process the fused multimodal input effectively, potentially leading to improved steering predictions.
\section{Experimental Setup}
\subsection{Dataset Description and Preparation}
Our experiments utilize the dataset described in \cite{Lechner2020}, which comprises RGB images captured from a vehicle-mounted camera.
The dataset consists of sequences of RGB images along with corresponding steering angle labels, capturing a wide variety of road types, weather conditions, and traffic scenarios, making it particularly suitable for our task. 
To focus on the most relevant parts of the scene for steering estimation, we apply preprocessing steps to each image. Following the procedure outlined in \cite{Lechner2020}, each image is cropped to dimensions of 78 × 200 pixels, excluding pixels corresponding to the sky and the front view of the car, and concentrating on the road and immediate surroundings. Pixel values are then normalized to ensure consistent input scaling across different lighting conditions.
We employ a 10-fold cross-validation approach to ensure robust evaluation of our models. The dataset is divided into 10 folds, denoted as [F.1, F.2, ..., F.10]. In each experimental run, one fold serves as the test dataset, while the remaining nine folds are used for training and validation. Within each training set, 10\% of the images are reserved for validation, with the remaining 90\% used for actual training. For example, when evaluating on fold F.1, 90\% of [F.2, F.3, ..., F.10] is used for training, 10\% for validation, and F.1 serves as the test set.
For each preprocessed RGB image, we extract additional modalities as described in the Modality Extraction section. Depth maps are generated using both MiDaS \cite{ranftl2020towards} and Monodepth2 \cite{godard2019digging} algorithms, while optical flow is computed between consecutive frames using the Farneback algorithm. These extracted modalities are aligned with their corresponding RGB images to create the multimodal input for our models.
This comprehensive dataset preparation process ensures that our models are trained and evaluated on a diverse and representative set of driving scenarios, allowing for a thorough assessment of our multimodal fusion approach to steering estimation. The use of multiple modalities derived from a single camera source enables us to explore the potential benefits of integrating spatial and temporal information for improved steering prediction accuracy.
\subsection{Implementation Details}
Our implementation focuses on efficiently processing multimodal data for steering estimation. Key details include:
\begin{itemize}
 \item \textit{Hardware and Software:} All experiments were conducted on NVIDIA GeForce RTX 3090 GPUs. We used Tensorflow as our deep learning framework, along with NumPy and OpenCV for data processing and optical flow computation.

 \item \textit{Training Procedure:} We train the models for 100 steps and employ the Adam optimizer for optimization. The batch size is fixed at 20, and we use a sequence length of 16 for the recurrent networks.

 \item \textit{Loss Function Components:} For VAE models, we use $\lambda_1 = 0.15$ and $\lambda_2 = \lambda_1 e^{-2}$ as regularization parameters for the reconstruction loss and KL divergence, respectively.

 \item \textit{Evaluation Metrics:} We use Mean Squared Error (MSE) and Mean Absolute Error (MAE) to evaluate the performance of our models. Additionally, we employ the Automatic Latent Perturbation (ALP) tool \cite{bairouk2024} for robustness analysis.
\end{itemize}
This implementation setup ensures reproducibility and comparability with existing state-of-the-art methods while allowing us to thoroughly explore the effectiveness of our multimodal fusion approach for steering estimation.
\section{Results \& Discussion}

\begin{table*}[!t]
    \centering
    \small
    \setlength{\tabcolsep}{3pt} 
    \begin{tabular}{  c |   c | c  c  c  c  c  c  c  c c  c | c | c }
        \multicolumn{1}{c}{} & \multicolumn{1}{c}{} & \multicolumn{10}{c}{MSE (per fold)} & \multicolumn{1}{c}{MSE} & \multicolumn{1}{c}{MAE}  \\
        \cline{3-14}
        \multicolumn{1}{c}{} & Model  & \rr{F.1} & \rr{F.2} & \rr{F.3} & \rr{F.4} & \rr{F.5} & \rr{F.6} & \rr{F.7} & \rr{F.8} & \rr{F.9}& \rr{F.10} & $\Delta \pm \delta$ & $\Delta \pm \delta$\\
        \hline
\multirow{18}{*}{\rotatebox{90}{Results from \cite{Lechner2020}}} & CNN                         & - & - & - & - & - & - & - & -   &-  &-  & 4.28$\pm$4.63  & - \\
                   & CNN-RNN                      & - & - & - & - & - & - & - & -   &-  &-  & 3.39$\pm$4.39  &  -\\
                   & CNN-LSTM (64 units)          & - & - & - & - & - & - & - & -   &-  &-  & 3.17 $\pm$3.85 &  - \\
                   & CNN-LSTM (19 units)          & - & - & - & - & - & - & - & -   &-  &-  & 3.38$\pm$4.48  &  \\
                   & CNN-Sparse LSTM (19 units)   & - & - & - & - & - & - & - & -   &-  &-  & 3.68$\pm$5.21  & - \\
                   & CNN-Sparse LSTM (64 units)   & - & - & - & - & - & - & - & -   &-  &-  & 3.25$\pm$3.93  &  -\\
                   & CNN-NCP (19 units)           & - & - & - & - & - & - & - & -   &-  &-  & 3.22$\pm$3.92  &  - \\
                   & CNN-NCP (randomly wired)     & - & - & - & - & - & - & - & -   &-  &-  & 5.19$\pm$5.43  &  - \\
                   & CNN-NCP (fully connected)    & - & - & - & - & - & - & - & -   &-  &-  & 5.18$\pm$4.19  &  \\
                   & VAE-NCP (19 units)           & - & - & - & - & - & - & - & -   &-  &-  & 4.67$\pm$3.74  &  - \\
                   & VAE-LSTM (64 units)          & - & - & - & - & - & - & - & -   &-  &-  & 4.70$\pm$4.80  &  - \\
                   & VAE-LSTM (19 units)		  & - & - & - & - & - & - & - & -  & - & - & 6.75$\pm$8.33  & - \\ 
                   \hline      
\multirow{4}{*}{I} & CNN-NCP using RGB & 0.64  & 0.80  & 2.14  & 17.12 & 0.97  & 1.17  & 4.72  & 1.24  & 5.81  & 2.21  & 3.68 $\pm$ 5.03& 0.81 $\pm$ 0.33 \\
& CNN-NCP using RGB$\mathrm{D}_1$ with EF & 0.55 & 0.73 & 2.54 & 26.92 & 1.14 & 1.35 & 7.56  & 2.17 & 5.53 & 2.68 & 5.12 $\pm$ 7.97& 0.91 $\pm$ 0.41\\     
& CNN-NCP using RGB$\mathrm{D}_1$ with HF & 0.63 & 0.79 & 1.71 & 19.14 & 0.95 & 1.35 & 6.39 & 1.6 & 6.16 & 2.65 & 4.13 $\pm$ 5.68 & 0.87 $\pm$ 0.38\\    
& CNN-NCP using RGB$\mathrm{D}_2$ with HF & 0.67 & 0.89 & 1.96 & 17.47 & 1.17 & 1.53 & 4.51 & 2.49 & 6.99 & 3.11 & 4.08 $\pm$ 5.08 & 0.89 $\pm$ 0.33 \\
& CNN-NCP using RGB-OF {\textbf{(Ours)}}&  
 \textbf{0.32} & \textbf{0.41} & \textbf{0.67} & \textbf{9.66} & \textbf{0.37} & \textbf{0.64} & \textbf{1.560} & \textbf{0.56} & \textbf{0.94} & \textbf{1.27} & \textbf{1.64} $\pm$ \textbf{2.84} & \textbf{0.56} $\pm$ \textbf{0.21}  \\
 \hline
\multirow{2}{*}{II} & VAE-NCP  using RGB & 0.74 & 1.48 & 3.03 & 23.98 & 2.41 & 3.00 & 7.29 & 3.17 & 7.76 & 3.97 & 5.68 $\pm$ 6.81 & 1.13 $\pm$ 0.40  \\
& VAE-NCP using RGB-OF {\textbf{(Ours)}}  & \textbf{0.41} & \textbf{0.60} & \textbf{0.78} & \textbf{9.23} & \textbf{0.52} & \textbf{0.94} & \textbf{1.49} & \textbf{0.85} & \textbf{2.15} & \textbf{1.27} & \textbf{1.82} $\pm$ \textbf{2.65} & \textbf{0.65} $\pm$ \textbf{0.22}  \\
 \hline
\multirow{2}{*}{III} & VAE-LSTM using RGB & 0.98 & 1.42 & 2.39 & 21.43 & 1.09 & 2.6   & 11.02 & 2.08 & 8.46 & 3.9 & 5.54 $\pm$ 6.51&  1.13 $\pm$ 0.45 \\
& VAE-LSTM using RGB-OF {\textbf{(Ours)}} & \textbf{0.24}  & \textbf{0.70} & \textbf{0.71} & \textbf{8.67} & \textbf{0.43} & \textbf{0.71} & \textbf{1.28} &  \textbf{0.67}     &  \textbf{3.30}     &  \textbf{1.07} & \textbf{1.77}  $\pm$ \textbf{2.57} & \textbf{0.59} $\pm$ \textbf{0.19} \\
\end{tabular}
    \caption{ Comparative evaluation of vehicle steering estimation error using different models with various feature fusion. $\mathrm{D}_1$ corresponds to the depth extracted using MiDas approach, while $\mathrm{D}_2$ corresponds to the one extracted using Monodepth2. EF and HF correspond to early  and late fusion respectively.}
    \label{tab:comparison}
\end{table*}

\begin{table}[!t]
    \centering
    \begin{tabular}{c | c }
    Features & SSIM  $(\Delta \pm \delta)$ \\
    \hline
        RGB-Depth (MiDas) & 0.3 $\pm$ 0.12 \\
        RGB-Depth (Monodepth2) &  0.26 $\pm$ 0.44 \\
        RGB-Optical Flow & 0.11 $\pm$ 0.036  
    \end{tabular}
    \caption{ Average of SSI comparison between RGB and Depth, and RGB and Optical Flow.}
    \label{tab:ssid}
\end{table}

\subsection{Results}

\begin{figure}[!t]
    \centering    \includegraphics[width=0.43\textwidth, keepaspectratio]{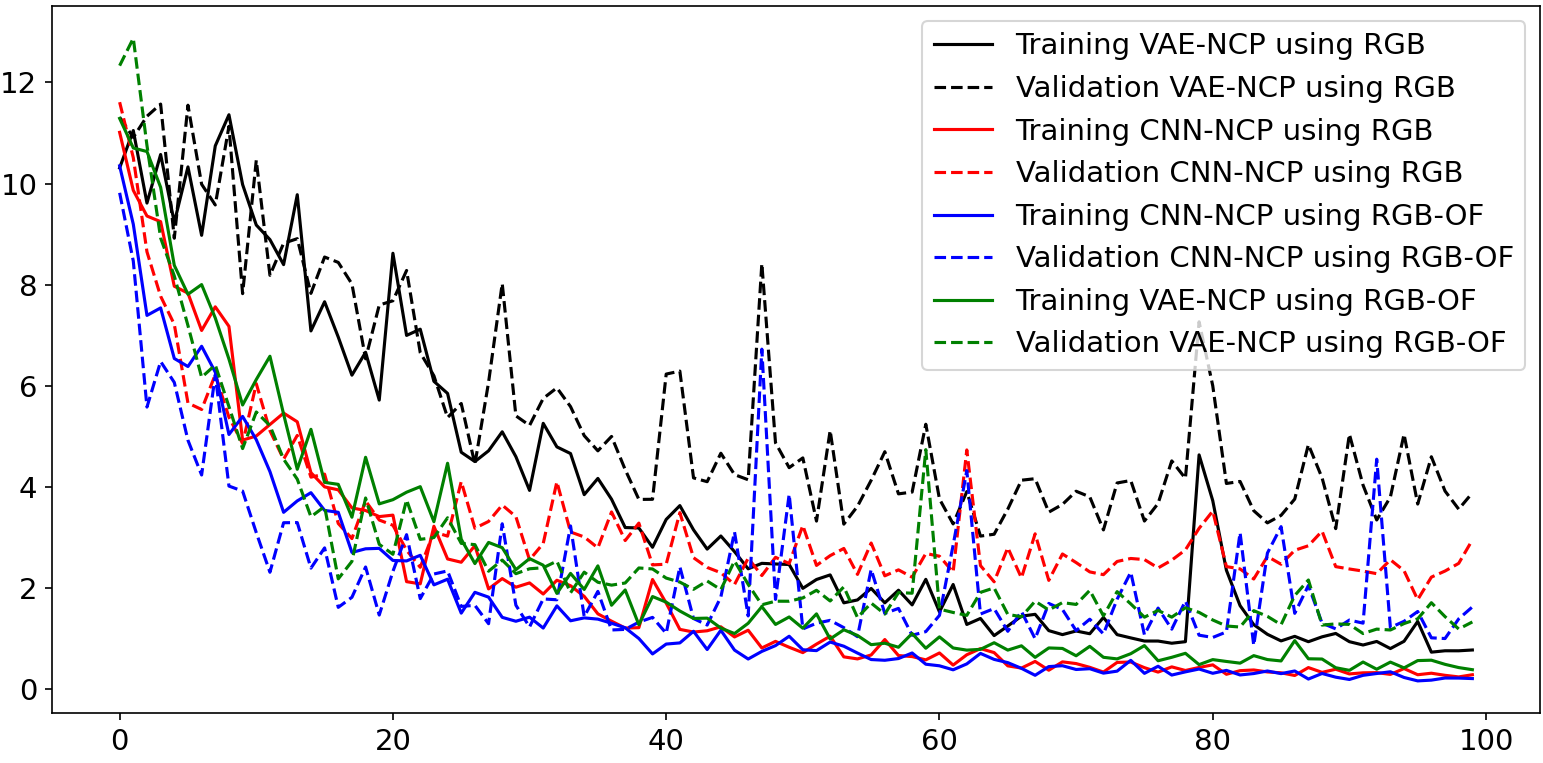} 
    \caption{Variation of MSE for training and validation across 100 steps when using either RGB and RGB-OF for VAE-NCP (black and green lines) and CNN-NCP models (red and blue lines), corresponding to F.1 in Table~\ref{tab:comparison}~(I \& II).}
    \label{fig:MSE_variation}
\end{figure}

Table \ref{tab:comparison} presents a comprehensive comparison of our proposed methods against different approaches and various modality combinations.

By analyzing category (I) from Table~\ref{tab:comparison}, which corresponds to the results obtained from the CNN-NCP model, and comparing the steering estimation using the RGB images and the depth extracted with the MiDas approach (RGB$\mathrm{D}_1$) with either EF or HF, we  notice that HF yielded lower Mean Squared Error (MSE) and lower standard deviation (4.13; 5.68) compared to EF (5.12; 7.97). This observation could be attributed to potential inaccuracies and misalignments between depth data and RGB images, which might confuse or distort the model predictions. Therefore, we consider the HF when using Monodepth2 (RGB$\mathrm{D}_2$), and we obtain similar results to RGB$\mathrm{D}_1$.
Interestingly, our experiments with depth modality (RGBD) did not yield improvements over RGB-only models. Therefore, we do not consider depth as an additional modality for the other models. However, early fusing optical flow with RGB (RGB-OF) enhances the model performance by incorporating motion-related data into the RGB color information, potentially offering valuable insights for vehicle steering estimation.  
This improvement is evident when comparing the fifth line to the other lines in category (I), where the average MSE and MAE are reduced by 31\%. Moreover, this enhancement is consistent across each fold, indicating consistent improvement.
The difference between the results obtained from adding depth and optical flow to the RGB images is likely due to the high correlation between RGB and depth information when extracted from a single camera source, as evidenced by the Structural Similarity Index (SSIM) values presented in Table \ref{tab:ssid}. The SSIM between RGB and depth maps (0.3 ± 0.12 for MiDaS and 0.26 ± 0.44 for Monodepth2) is significantly higher than that between RGB and optical flow (0.11 ± 0.036), indicating that optical flow provides more complementary information to RGB.

Encouraged by these results, we seek to further investigate this approach by applying it to two other models, VAE-NCP and VAE-LSTM. The results presented in Table~\ref{tab:comparison} (II and III) show the significant decrease in MSE and MAE. Specifically, MSE is reduced by 68\%, and MAE by more than 40\% compared to using solely RGB images. This enhancement extends beyond steering estimation. As highlighted in \cite{bairouk2024}, VAE-NCP offers superior interpretability compared to CNN-NCP, albeit at the cost of decreased accuracy in terms of MSE (3.68; 5.68) when using only RGB images. However, with our proposed approach, both CNN-NCP and VAE-NCP methods achieve nearly identical improved MSE (1.64; 1.82), indicating that we have attained a model with enhanced interpretability while maintaining the accuracy level of improved CNN-NCP.

The superiority of the RGB-OF approach is further illustrated in Figure~\ref{fig:MSE_variation}, which shows the variation of MSE for training and validation across 100 steps. The RGB-OF models (green and blue lines) demonstrate consistently lower error rates compared to their RGB-only counterparts (black and red lines), particularly on the validation set. This suggests that the inclusion of optical flow not only improves overall performance but also enhances the model ability to generalize to unseen data.


In summary, these results strongly support our hypothesis that optical flow provides valuable complementary information to RGB data for steering estimation. The significant and consistent performance improvements across various architectures and evaluation metrics demonstrate the potential of our approach to enhance the accuracy and robustness of autonomous driving systems.

\subsection{Robustness Analysis using ALP}

\begin{figure*}[!t]
    \centering    
    \includegraphics[width=0.67\textwidth]{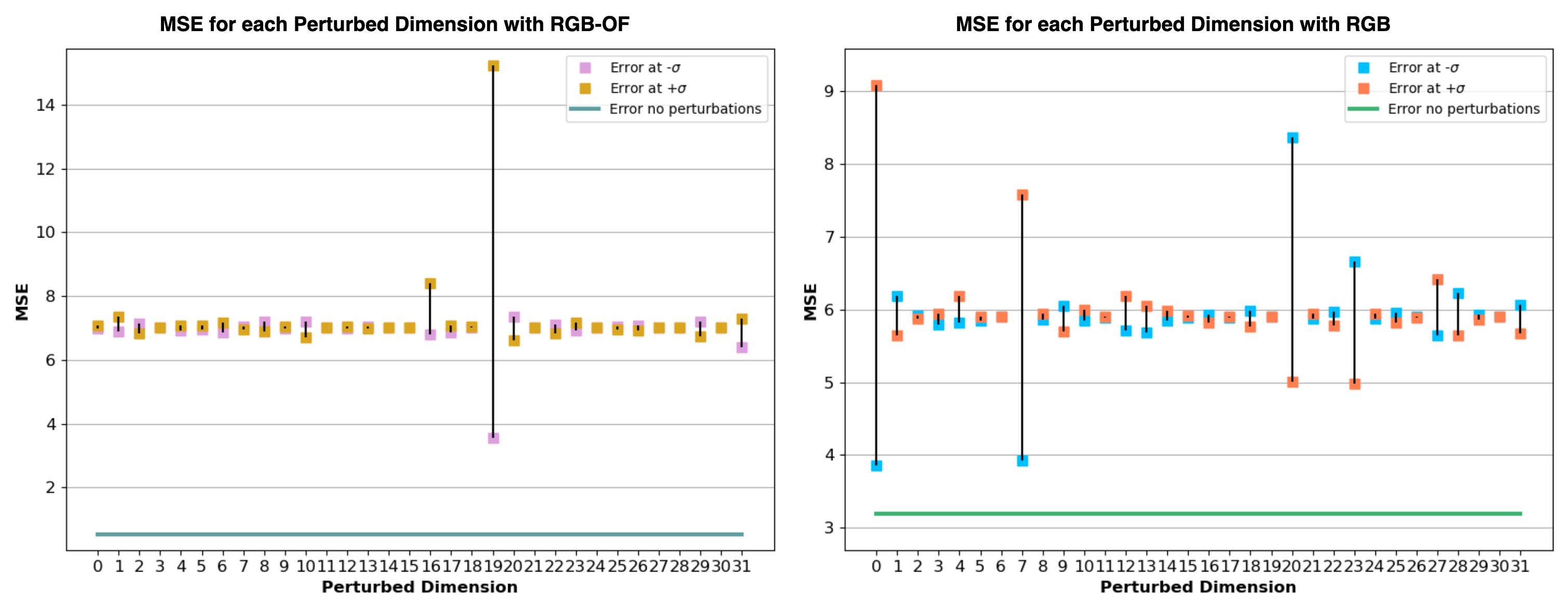} 
    \caption{
    Steering Error Response to Latent Dimension Perturbations: the left plot depicts the MSE variability under perturbation of each latent dimension in a model using RGB and optical flow, while the right plot displays the variability in MSE when each latent dimension is perturbed in a model using only RGB images. 
    The perturbation value was set to $\sigma=0.3$.}
    \label{fig:MSE}
\end{figure*}

\begin{figure*}[!t]
    \centering    \includegraphics[width=0.67\textwidth]{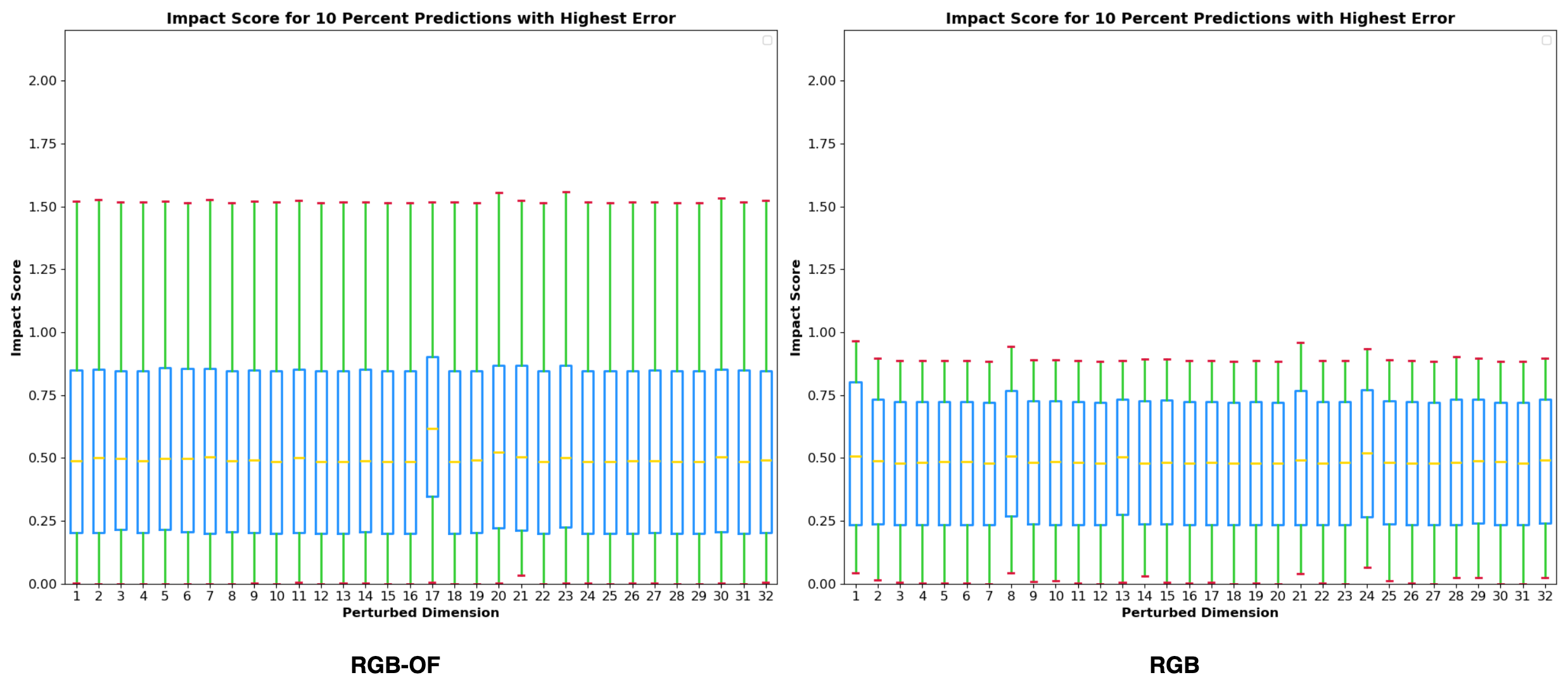} 
    \caption{Impact Score Distribution for the Top 10\% of Predictions with Highest Error: 
    The left plot corresponds to the RGB-OF model, illustrating a consistent impact score across perturbed dimensions, while the right plot corresponds to  the RGB-only model,  displaying a more varied impact score.
    }
    \label{fig:ImpactScore}
\end{figure*}

To further analyze the robustness of the proposed approach, we employed the Automatic Latent Perturbation (ALP) tool proposed in\cite{bairouk2024}. This tool enhances the analysis of model robustness and interpretation of latent dimensions in VAEs, particularly useful in high-dimensional spaces. Traditional methods, which manually perturb latent variables to analyze changes in outputs, suffer from scalability issues and limited effectiveness in complex models. ALP automates this by perturbing each latent dimension and generating both positive and negative reconstructions, allowing for an assessment of the specific information each dimension encodes.

Moreover, the ALP has been extended to evaluate the effects of latent dimension perturbations on operational outputs like steering commands in autonomous vehicles. This involves generating steering predictions by perturbing the latent vectors and passing them through a secondary model component, like the NCP in our case. By comparing the steering results from different perturbations, an impact score is computed, quantifying the influence of each latent dimension on steering behavior. The impact score formula is as follows:
\begin{equation}
\text{I}_j = \frac{|\hat{\mathbf{y}}^{j,+} - \hat{\mathbf{y}}^{j,0}| + |\hat{\mathbf{y}}^{j,-} - \hat{\mathbf{y}}^{j,0}| + |\hat{\mathbf{y}}^{j,+} - \hat{\mathbf{y}}^{j,-}|}{3}
\end{equation}
where $\mathbf{I}$ is the impact score, $\hat{\mathbf{y}}^{j,+}$ and $\hat{\mathbf{y}}^{j,-}$ are the steering predictions for the perturbed latent vectors increased and decreased by a perturbation value of $\sigma = 0.3$ along the $j$-th dimension, respectively, for both RGB and RGB-OF setups, and $\hat{\mathbf{y}}^{j,0}$ is the steering prediction for the unperturbed latent vector.

We conduct experiments on fold F.$2$, utilizing the complete fold to ensure a comprehensive analysis. 
Figure~\ref{fig:MSE} depicts how the incorporation of optical flow with RGB images affects the model MSE under both perturbed and unperturbed conditions. 
This figure provides insights into the variability and stability of the model predictions across different perturbed dimensions.  Figure~\ref{fig:MSE} confirms, as previously mentioned, that integrating optical flow with RGB images enhances model predictions for steering control in unperturbed conditions, compared to using only RGB images. Furthermore, when analyzing the latent space perturbations, it is evident from Figure~\ref{fig:MSE} that the model using both RGB and optical flow manifests significantly less variation in MSE across perturbed dimensions compared to the RGB-only model. This reduced variation in response to perturbations within the latent space highlights the model robustness; it shows more resilience to internal variability, which is crucial for dependable decision-making in dynamic environments. On the other hand, the higher variability across dimensions in the model based only on RGB images (see right plot in Figure~\ref{fig:MSE}) indicates less stable and reliable predictive behavior when the latent space is disturbed.

Furthermore, Figure~\ref{fig:ImpactScore} presents the distribution of impact scores for the top 10\% of steering predictions that exhibit the highest error, providing a direct measure of the model sensitivity to deviations in input data.
We note a consistently higher impact score across the 32 perturbed dimensions for RGB-OF (left plot) in comparison to utilizing only RGB (right plot). This indicates that the RGB-OF model is potentially more adept at identifying when inputs or situations deviate from the norm, which could lead to better handling of Out-Of-Distribution (OOD) scenarios. The RGB model, with lower impact scores, might be less sensitive to these perturbations, potentially making it less effective at OOD detection. Consequently, the RGB-OF model can be considered more robust in this specific aspect of uncertainty identification and OOD situation handling. This increased sensitivity can be an asset in applications like autonomous driving systems, where recognizing OOD situations that differ from the training data is crucial for safety and performance.

\section{Conclusion}
This study introduces a novel approach to enhance end-to-end steering estimation for autonomous vehicles by leveraging multimodal information extracted from a single monocular camera. Our primary contribution lies in demonstrating the significant impact of incorporating optical flow information alongside RGB imagery. The fusion of RGB and optical flow data consistently outperforms state-of-the-art methods and other modality combinations, achieving a remarkable 31\% reduction in steering estimation error. This improvement is consistent across various neural network architectures, including CNN-NCP, VAE-NCP, and VAE-LSTM, highlighting the robustness and versatility of our approach. Moreover, the integration of optical flow enhances the model's ability to generalize to unseen data, as evidenced by lower error rates on validation sets.

These findings have important implications for the development of autonomous driving systems. By demonstrating that rich, multimodal information can be extracted and effectively utilized from a single camera source, our approach offers a cost-effective and computationally efficient solution for improving steering estimation accuracy. Future work could explore the integration of our approach with other sensor modalities and its application to other autonomous driving tasks. In conclusion, our research underscores the importance of leveraging complementary information sources in autonomous driving systems, potentially enhancing the safety and reliability of self-driving vehicles.
{\small
\bibliographystyle{ieee_fullname}
\bibliography{egbib}
}

\end{document}